# Otimização de pesos e funções de ativação de redes neurais aplicadas na previsão de series temporais

Gecynalda S. da S. Gomes        Teresa B. Ludermir

*Resumo*: Redes Neurais tem sido aplicada para previsao de series temporais com bons resultados experimentais que indicam a alta capacidade de aproximaçao de funções com boa precisão. A maioria dos modelos de redes neurais utilizados nestas aplicações utilizam funçoes de ativação com parâmetros fixos. Entretanto, é sabido que a escolha da função de ativação influencia fortemente a complexidade e o desempenho da rede neural e que um número limitado de funções de ativação tem sido utilizado. Neste trabalho, propomos a utilizaçao de uma família de funções de ativação assimétrica  de parâmetro livre para redes neurais e mostramos que essa família de funções de ativação definida satisfaz os requisitos do teorema da aproximação universal. Uma metodologia para a otimizaçao global dessa família de funções de ativação com parâmetro livre e dos pesos das conexões entre as unidades de processamento da rede neural é utilizada. A ideia central da metodologia proposta é otimizar simultaneamente os pesos e a função  de ativação usada em uma rede multilayer perceptron (MLP), através de uma abordagem que combina as vantagens de *simulated annealing*, de *tabu search* e de um algoritmo de aprendizagem local, com a finalidade de melhorar o desempenho no ajuste e na previsao de séries temporais. Escolhemos dois algoritmos de aprendizagem: o *backpropagation* com o termo *momentum* (BPM) e o LevenbergMarquardt (LM).

*Abstract*: Neural Networks have been applied for time series prediction with good experimental results that indicate the high capacity to approximate functions with good precision. Most neural models used in these applications use activation functions with fixed parameters. However, it is known that the choice of activation function strongly influences the complexity and performance of the neural network and that a limited number of activation functions have been used. In this work, we propose the use of a family of free parameter asymmetric activation functions for neural networks and show that this family of defined activation functions satisfies the requirements of the universal approximation theorem. A methodology for the global optimization of this family of activation functions with free parameter and the weights of the connections between the processing units of the neural network is used. The central idea of the proposed methodology is to simultaneously optimize the weights and the activation function used in a multilayer perceptron network (MLP), through an approach that combines the advantages of simulated annealing, tabu search and a local learning algorithm, with the purpose of improving performance in the adjustment and forecasting of time series. We chose two learning algorithms: backpropagation with the term momentum (BPM) and LevenbergMarquardt (LM).

## I. Introdução

Modelos não-lineares de redes neurais artificiais (RNA) fazem parte de uma importante classe de modelos que tem atraído atenção considerável em muitas aplicações. O uso desses modelos em muitos trabalhos aplicados é, muitas vezes, motivado por resultados empíricos indicando que, sob condições  de regularidade, modelos relativamente simples de RNAs são capazes de aproximar qualquer função mensurável de Borel a qualquer grau de decisão [1]. Redes neurais com uma simples camada escondida usando funções sigmóides são aproximadores universais de funções, ou seja, estes modelos podem aproximar funções contínuas arbitrárias dado um número suficiente de neurônios [1], [2], [3], [4].

Mapeamentos de entrada-saída podem ser representados por redes neurais através da combinação de conexões ponderadas entre os neurônios da rede [4]. Funahashi [3] provou que qualquer mapeamento contínuo pode ser realizado por uma rede *multilayer perceptron* (MLP) com funções  de ativação diferenciáveis e monotonicamente crescentes, em seu trabalho foi usado funções  sigmóides na camada escondida e na camada de saída funções lineares.

No problema de previsão de séries temporais, há várias décadas, muitos autores vêm utilizando diferentes métodos estatísticos para modelagem e previsão que variam de médias móveis e alisamento exponencial a regressões lineares ou não-lineares. Box e Jenkins [5] desenvolveram os modelos autorregressivos integrados médias móveis (ARIMA) para prever séries temporais. Para melhorar previsões de séries temporais com características não-lineares, vários pesquisadores desenvolveram métodos alternativos que modelam essas aproximações, por exemplo, modelos autorregressivos heteroscedásticos (ARCH) [6]. Apesar destes métodos mostrarem melhorias significativas sobre os modelos lineares, eles tendem a ser específicos para determinadas aplicações. Como os modelos de RNAs são usados como aproximadores universais de funções [1], muitos pesquisadores os vêm utilizando para prever diversos eventos não-lineares de séries temporais para avaliar a eficácia do desempenho desses modelos em relação aos modelos tradicionais de previsão [7], [8], [9], [10].

Em geral, o desempenho de RNAs depende do número de camadas escondidas, do número de neurônios escondidos, do algoritmo de aprendizagem e da função de ativação de cada neurônio. Entretanto, a maioria dos trabalhos relacionados com redes neurais está associada com os algoritmos de aprendizagem e seleção de arquitetura, negligenciando a importância das funções de ativação. A escolha da função de ativação pode influenciar fortemente a complexidade e o desempenho de RNA, além de ter um importante papel na convergência do algoritmo de aprendizagem [11], [12], [13], [14], [15].

Vários tipos de funções de ativação foram propostos. Pao [16] utilizou uma combinação de várias funções (polinomial, periódica, sigmóide e Gaussiana). Hartman *et al.* [17] propuseram funções gaussianas como funções de ativação na camada escondida como aproximadores universais de funções. Hornik [18], [19] e Leshno *et al.* [20] utilizaram funções de ativação não-polinomiais. Leung e Haykin [21] usaram funções racionais com ótimos resultados. Giraud *et al.* [22] usaram a função *Lorentzian*. Rosen-Zvi et al. [23] mostraram resultados gerais de modelos de redes neurais com funções de ativação periódicas. Skoundrianos *et al.* [24] propuseram uma nova função sigmóide com bons resultados para modelagem de sistemas de tempo dinâmicos e discretos. Ma e Khorasani [25] usaram como função de ativação a polinomial Hermite. Gomes e Ludermir [26] propuseram o uso de duas novas funções de ativação, complemento log-log e probit, que apresentaram bons desempenhos em relação a função de ativação logit. Elas também mostraram que essas funções de ativação são aproximadores universais e que são adequadas para problemas de regressão. Uma característica comum nessas funções de ativação é que elas são todas de parâmetros fixos e não podem ser ajustadas para resolver os diferentes tipos de problemas.

Existem poucos trabalhos com ênfase em funções de ativação com parâmetros livres. Alguns estudos têm mostrado que redes neurais com funções de ativação de parâmetros livres apresentaram melhores resultados do que redes com arquiteturas clássicas cuja função de ativação tem parâmetros fixos. Em Chen e Chang [27], variáveis de ganho e de inclinação na função de ativação sigmóide generalizada proposta são ajustadas durante o processo de aprendizagem mostrando melhoria na modelagem dos dados. Guarnieri *et al.* [28] apresentaram uma nova função de ativação *spline* adaptavivo, estudaram suas propriedades e mostraram uma melhoria tanto na complexidade quanto no desempenho da rede neural em termos de capacidade de generalização. Singh e Chandra [14] propuseram uma nova classe de funções sigmóides, provaram que a função do envelope das derivadas da classe definida também é uma sigmóide e mostraram que essas funções satisfazem os requisitos do teorema da aproximação universal. Chandra [29] propõe dois métodos de parametrização que permitem construir classes sigmóides baseadas em qualquer sigmóide dada e demonstra que todos os membros das classes propostas satisfazem os requisitos para serem utilizadas como função de ativação em redes neurais. Gomes *et al.* [30] usaram como função de ativação com parâmetro livre uma função baseada na transformação Aranda-Ordaz assimétrica [31] obtendo bons resultados para aproximação de funções de regressão, tanto com o uso do algoritmo *backpropagation* como com o uso do algoritmo Levenberg-Marquardt. Estes e outros artigos vistos mostram que a escolha da função de ativação é considerada por muitos especialistas tão importante quanto a arquitetura e o algoritmo de aprendizagem da rede neural.

A função sigmóide logística assume um intervalo contínuo de valores entre 0 e 1. Quando é desejável que a função de ativação se estenda de −1 a +1, assumindo uma forma antisimétrica em relação a origem utiliza-se, geralmente, a função tangente hiperbólica, caso queira manter a característica de uma função sigmóide [32]. Porém, quando a probabilidade de uma dada resposta se aproxima de 0 a uma taxa diferente da que se aproxima de 1, funções simétricas são inapropriadas [33]. Com base nestes fatos, fica a pergunta com relação as características da função de ativação na determinação das propriedades do processo de aprendizagem da rede neural: Qual é a relevância da simetria nas funções de ativação?

Para responder a esta pergunta, neste trabalho, propomos a utilização de uma família de funções de ativação assimétricas com parâmetro livre (FFAAPL) para redes neurais baseada na família de transformações Aranda-Ordaz com aspectos assimétricos [31]. Essa família de funções Aranda-Ordaz foi proposta para dados binários, e é utilizada como funções de ligação em modelos lineares generalizados (MLG) quando os dados seguem uma distribuição binomial. Para maiores detalhes ver [34], [35].

Para otimizar o valor do parâmetro da FFAAPL, bem como os pesos e bias da rede neural, usamos o método de otimização global *simulated annealing* juntamente com *tabu search*. Este método pode ser combinado com uma técnica baseada em gradiente (eg, o algoritmo *backpropagation*) em uma abordagem de treinamento híbrido agregando a eficiência dos métodos de otimização global com o ajuste fino das técnicas baseadas em gradiente. Esta abordagem é baseada nos trabalhos de Ludermir *et al.* [36] e Carvalho e Ludermir [37], [38], porém, nesses trabalhos são feitas otimizações globais de pesos e arquitetura da rede, usando a função tangente hiperbólica com parâmetro fixo como função de ativação em todos os problemas abordados. No nosso trabalho, vamos fixar a topologia da rede para que sejam avaliados os reais efeitos da otimização dos pesos juntamente com a função de ativação com parâmetro livre.

Geralmente, os modelos existentes de RNAs para previsão usam redes MLP, em que a quantidade de camadas escondidas, a quantidade de nodos das camadas de entrada e escondidas e a função de ativação são escolhidos, frequentemente, por tentativa e erro com a finalidade de encontrar um modelo plausível para a aplicação específica. Ghiassi e Saidane [39] desenvolveram um modelo de rede neural - DAN2: Uma arquitetura dinâmica para RNAs - que emprega uma arquitetura diferente dos modelos tradicionais. Para demonstrar a eficácia do modelo DAN2, os autores compararam o seu desempenho com os desempenhos dos modelos tradicionais de RNAs e ARIMA em séries não-lineares mostrando superioridade do modelo proposto por eles para o ajuste e previsão de séries temporais. Por este motivo, implementamos o modelo DAN2 para servir de referência e compararmos com os resultados do modelo de rede neural com a FFAAPL proposta.

Portanto, a ideia é encontrar um modelo de redes neurais MLP que tenha bom desempenho no ajuste e na previsão de séries temporais capaz de modelar séries temporais cujo comportamento seja o mais variado possível. Este modelo combina as técnicas de *simulated annealing*, *tabu search* e um algoritmo de aprendizagem, *backpropagation* com o termo *momentum* (BPM) ou Levenberg-Marquardt (LM), cuja função de ativação tem parâmetro livre e a arquitetura da rede contém uma camada escondida e poucos nodos escondidos. Com isso, o objetivo é proporcionar uma maior estabilidade nos resultados de previsão de séries temporais. A família de funções de ativação a ser utilizada tem como casos especiais a função logit e a função complemento log-log [31] e satisfaz os requisitos do teorema da aproximação universal.

Este artigo está organizado da seguinte forma: na Seção II apresentamos os trabalhos relacionados com a otimização global e redes neurais, na Seção III apresentamos a prova matemática em que as novas funções satisfazem o teorema da aproximação universal, a metodologia de otimização está apresentada na Seção IV, na Seção V, apresentamos as configurações experimentais e os resultados. Por fim, na Seção VI estão as considerações finais.

## II. OTIMIZAÇÃO GLOBAL DE REDES NEURAIS

Diversas técnicas de otimização vêm sendo usadas na literatura visando melhorar o desempenho de redes neurais artificiais, tais como *simulated annealing* (SA), *tabu search* (TS), algoritmos genéticos (AGs), entre outras. Essas técnicas, geralmente, são utilizadas como uma abordagem híbrida para treinamento da rede neural. Em geral, o objetivo é minimizar o problema principal do algoritmo baseado em gradiente: a convergência local.

Uma integração de SA, TS e AGs foi proposta por Liu *et al.* [40]. Em Li *et al.* [41], AGs e SA foram combinados para otimização de processos em planejamentos de engenharia. Em geral, sabe-se que técnicas de otimização global, como SA e TS, são relativamente ineficientes para ajuste fino em buscas locais. Dessa forma, é importante investigar se o desempenho de generalização das redes ainda pode ser melhorado quando as topologias geradas por estas técnicas são treinadas com uma abordagem de busca local, como o algoritmo *backpropagation*. Esta combinação de otimização global com técnicas locais foi utilizada por Yao [42] em trabalhos com AGs. No treinamento de RNAs, essas misturas de técnicas foram utilizadas em diversas aplicações de forma simultânea ou não [36], [43], [44], [45].

Tsai *et al.* [46] utilizaram um algoritmo híbrido para o ajuste da arquitetura e parâmetros de redes neurais artificiais *feedforward*. Ferreira e Ludermir [47] utilizaram um processo algoritmos genéticos para otimização de reservoir computing. O método de SA foi usado com sucesso em alguns problemas de otimização global, como pode ser visto em Corana *et al.* [48]. Porto *et al.* [49] implementaram SA e *backpropagation* para treinamento de uma rede MLP com topologia fixa contendo duas camadas ocultas, o problema abordado foi o reconhecimento de respostas de sonar. Sexton *et al.* [43] usaram SA e AG, cujas soluções candidatas foram representadas por vetores de números reais contendo todos os pesos da rede. Em Hamm [50] SA foi usado para otimizar pesos de redes neurais artificiais. Yamazaki e Ludermir [44] usam SA e TS de forma simultânea para otimizar pesos e arquitetura, neste caso o problema considerado foi o reconhecimento de odor em um nariz artificial. Ludermir *et al.* [36] combinam três técnicas, SA, TS e o algoritmo de treinamento *backpropagation*, para gerar um processo automático para produzir redes com bom desempenho de classificação e baixa complexidade. Zanchettin e Ludermir [45] apresentam um método de otimização que integra quatro técnicas, SA, TS, AG e o algoritmo de treinamento *backpropagation* para encontrar pesos e arquitetura de uma rede neural.

Na maioria das abordagens, os autores utilizam essas técnicas para otimizar os parâmetros e valores iniciais para conexões de peso entre as unidades de processamento e arquitetura da rede, fixando uma função de ativação comumente usada na literatura como sigmóide logística ou tangente hiperbólica. Por exemplo, em [44], [36], [45] a função de ativação utilizada em todos os problemas foi a tangente hiperbólica com parâmetro fixo. Existem trabalhos em que se utilizam funções de ativação com parâmetro livre, porém, alguns autores usam como metodologia de busca do melhor valor do parâmetro uma adaptação do algoritmo *backpropagation* [27], [51], [15], sendo que esse tipo de abordagem continua enfrentando o problema da otimização local. Outros autores utilizam métodos de otimização semelhante ao *line search* [30]. Assim, surge a ideia de otimizar simultaneamente a função de ativação e os pesos da rede.

## III. FAMÍLIA DE FUNÇÕES DE ATIVAÇÕES ASSIMÉTRICAS

Uma função sigmóide pode ser definida como [29]

**Definiçao 1:** Uma função real, $f(x)$, $f: \mathbb{R} \to \mathbb{R}$, com as propriedades

$$\lim_{x \to +\infty} f(x) = a; \qquad \lim_{x \to -\infty} f(x) = b, \qquad (1)$$

onde $a$ e $b$ são números reais e $a > b$. Os valores usuais são $a = 1$ e $b = 0$ ou $-1$.

A classe geral de funções sigmóides incluem funções descontínuas como a função de Heaviside, a função arco tangente, a função tangente hiperbólica e função log-sigmóide, entre outras. Qualquer função que é não-constante, limitada e monotonicamente crescente satisfaz a equação (1) e consequentemente pertence ao conjunto de todas as funções sigmóides. Para funções sigmóides, incluindo a família de funções sigmóides assimétricas ($F_a$), o teorema da aproximação universal (TAU) pode ser resumido como [32].

O TAU fornece a justificativa matemática para a aproximação de uma função contínua arbitrária em oposição a representação exata [32]. O TAU prove um conjunto de condições que uma função de ativação precisa satisfazer para ser usada em redes neurais. As condições exigidas para que o vetor de entrada faça parte de um hipercubo unitário pode ser estendido para qualquer hipercubo limitado. Para isso utiliza-se um algoritmo eficiente que requisite que a função de ativação seja diferenciável e satisfaça uma simples equação diferencial para avaliar o incremento dos pesos sinápticos da rede neural.

A função Aranda-Ordaz assimétrica é definida por

$$\eta = \log\left\{\frac{(1-\pi)^{-\lambda}}{\lambda}\right\},$$

onde $\pi \in (0,1)$ e $\lambda > 0$. Assim, obtendo a função inversa, temos,

$$\pi = 1 - (1 + \lambda \exp(\eta))^{\frac{-1}{\lambda}}. \quad (2)$$

Portanto, para modelos de redes neurais, utilizaremos a função (2) para que seja uma família de sigmóide assimétrica ($F_\lambda$). Logo, em nosso contexto, temos

$$f_\lambda(x) = 1 - (1 + \lambda e^x)^{-1/\lambda}, \qquad \lambda > 0. \quad (3)$$

A Figura 1 apresenta o comportamento de $f_\lambda(x)$ como função de $x$ para a função de ativação para diferentes valores de $\lambda$. Observa-se que as funções sigmóide logística e complemento log-log são casos especiais da família de sigmóides $F_\lambda$ quando $\lambda = 1$ e $\lambda \to 0$, respectivamente. Para valores de $\lambda > 1$, $f_\lambda(x)$ se aproxima mais lentamente de um do que na função sigmóide logística.

Para todo membro da família $F_\lambda$, as proposições a seguir estabelecem que esses membros são não-constantes, limitados e monotonicamente crescentes.

*Proposição 3.1:* Para todo membro da família $F_\lambda$ é uma função monotonicamente crescente (MC).

**Prova.** Seja $x_2 > x_1$, então, $e^{x_2} > e^{x_1}$, pois $\forall x, e^x > 0$, logo MC. Sabemos que $\lambda > 0$, logo $1 + \lambda e^{x_2} > 1 + \lambda e^{x_1}$ também é uma função MC. Logo, $(1 + \lambda e^{x_2})^{1/\lambda} > (1 + \lambda e^{x_1})^{1/\lambda}$. Invertendo as funções, temos $(1+\lambda e^{x_2})^{-1/\lambda} < (1+\lambda e^{x_1})^{-1/\lambda}$. Multiplicando por $-1$ em ambos os lados, temos $-(1 + \lambda e^{x_2})^{-1/\lambda} > -(1 + \lambda e^{x_1})^{-1/\lambda}$. Adicionando um constante positiva, temos $1 - (1 + \lambda e^{x_2})^{-1/\lambda} > 1 - (1 + \lambda e^{x_1})^{-1/\lambda}$. Portanto, todo membro $f_\lambda(x)$ é uma função MC.

*Proposição 3.2:* Todo membro $f_\lambda(x)$ da família $F_\lambda$ é limitada em 1 quando $x \to +\infty$ e em 0 quando $x \to -\infty$, ou seja, $\forall \lambda > 0$, as relações a seguir são verdadeiras:

$$\lim_{x \to +\infty} f_\lambda(x) = 1; \qquad \lim_{x \to -\infty} f_\lambda(x) = 0. \quad (4)$$

**Prova.** Para o limite superior, temos $\lim_{x \to +\infty} f_\lambda(x) = \lim_{x \to +\infty}(1 - (1 + \lambda e^x)^{-1/\lambda})$. Pelas propriedades de limite temos $\lim_{x \to +\infty} f_\lambda(x) = 1 - \lim_{x \to +\infty}((1 + \lambda e^x)^{-1/\lambda})$. Logo, $\forall \lambda > 0$, temos $\lim_{x \to +\infty} f_\lambda(x) = 1 - 0 = 1$. Para o limite inferior, temos $\lim_{x \to -\infty} f_\lambda(x) = \lim_{x \to -\infty}(1 - (1 + \lambda e^x)^{-1/\lambda})$. Pelas propriedades de limite temos $\lim_{x \to -\infty} f_\lambda(x) = 1 - \lim_{x \to -\infty}((1+\lambda e^x)^{-1/\lambda})$. Logo, $\forall \lambda > 0$, temos $\lim_{x \to -\infty} f_\lambda(x) = 1 - 1 = 0$.

*Proposição 3.3:* Todo membro da família $F_\lambda$ é uma função diferenciável e satisfaz uma equação diferencial.

**Prova.** Diferenciar a equação (3)

$$\frac{df_\lambda(x)}{dx} = (1 + \lambda e^x)^{-1/\lambda} e^x (1 + \lambda e^x)^{-1}$$
$$= (1 + \lambda e^x)^{-(1+\lambda)/\lambda} e^x \quad (5)$$

A partir das Proposições 1-3, temos que todo membro da família $F_\lambda$ é não-constante, limitado e monotonicamente crescente. Assim, todo membro da FFAAPL satisfaz as propriedades requeridas no TAU, logo, pode ser usado como função de ativação da rede neural.

## IV. METODOLOGIA DE OTIMIZAÇÃO

Otimizar é melhorar o que já existe, projetar o novo com mais eficiência e menor custo. A otimização visa determinar a melhor configuração de projeto sem ter que testar todas as possibilidades envolvidas. O processo de busca normalmente parte de uma solução inicial ou de um conjunto delas, realizando melhoramentos progressivos até chegar a um outro conjunto que contenha uma ou todas as melhores soluções possíveis dentro do espaço de busca. A solução de um problema de otimização pode ser caracterizada como um processo de busca local ou global.

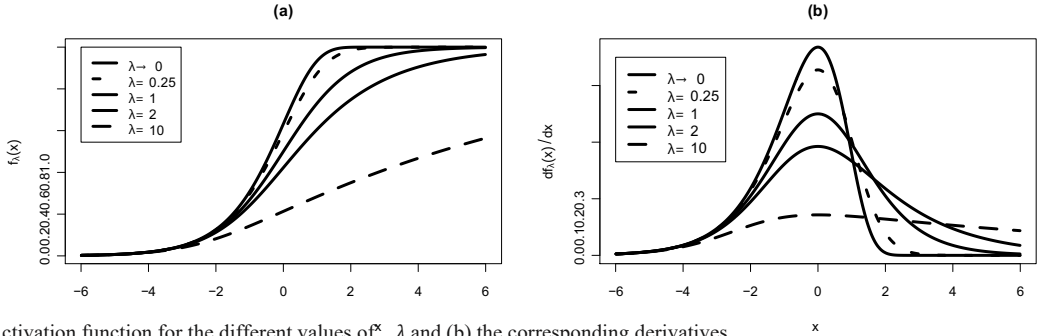

Figura 1. (a) Activation function for the different values of $\lambda$ and (b) the corresponding derivatives.

O processo de busca local objetiva encontrar a melhor solução dentro de uma conjunto de soluções em um espaço restrito, sendo que esta solução depende do ponto de início do processo de busca. No processo de busca global o objetivo é encontrar a melhor solução possível, independentemente das condições de início do processo de busca. Quando existe um universo enumerável de possíveis combinações de elementos que se pretende minimizar ou maximizar, tem-se uma classe de otimização combinatória que se caracterizam pela estratégia de busca empregada, utilização de informações sobre o domínio do problema e complexidade. Os AGs [52], [53], SA [54] e TS [55] são algoritmos iterativos que, em geral, servem para resolver problemas de otimização combinatória.

O algoritmo SA pode ser definido como uma técnica de busca global que aproxima o máximo (ou o mínimo) de uma função objetivo $f: S \to R$, sobre um conjunto finito $S$. O algoritmo foi introduzido na literatura por Kirkpatrick *et al.* [54], baseado nas ideias de Metropolis *et al.* [56] sobre simulação de um sistema de partículas ao experimentar mudanças em temperatura. Sob perturbação, o sistema tenta encontrar um ponto de equilíbrio que minimize a energia total. O termo *annealing* em termodinâmica se refere ao esfriamento de materiais sob condições controladas. Kirkpatrick *et al.* [54] fizeram uma analogia entre os estados do sistema no problema de Metropolis e as possíveis configurações num problema de otimização mais geral, com os valores da função objetivo fazendo o papel dos níveis de energia e a temperatura do sistema correspondendo a um parâmetro de controle no processo de otimização.

Para escapar de mínimos locais, o algoritmo SA se diferencia dos demais métodos de busca, citados anteriormente, por aceitar movimentos que caracterizam uma degradação em seu desempenho [57]. O processo de busca consiste de uma sequência de iterações. Cada iteração consiste em alterar aleatoriamente a solução atual para criar uma nova solução na sua vizinhança. Uma vez que uma nova solução é criada, a correspondente alteração na função de custo é computada para decidir se a nova solução pode ser aceita. Se o custo de uma nova solução é menor do que o custo da solução atual, a nova solução é aceita. Caso contrário, o critério de Metropolis é verificado [56], com base na probabilidade de Boltzmann. Esta probabilidade é regulada por um parâmetro chamado temperatura, que decresce durante o processo de otimização. Assim, o parâmetro $T$ é referenciado como a temperatura e o processo da redução da temperatura é chamado de processo de resfriamento.

Neste trabalho, a estratégia de resfriamento escolhida é a regra de arrefecimento logarítmica Belisle obtida em [58]. De acordo com esta regra, a nova temperatura igual a temperatura atual, multiplicado por um factor de redução determinado dado por

$$\frac{1}{\log([(i-1)/I_T] \times I_T + \exp(1))} \quad (6)$$

onde $[a]$ representa a parte inteira da divisão. A temperatura inicial $T_0$, o número de funções avaliadas a cada temperatura, $I_T$, e o número máximo de iterações, $I_{max}$, são parâmetros da implementação. Em muitas situações, o método de SA pode apresentar certa lentidão na convergência para soluções aceitáveis, dependendo do esquema de esfriamento. Se a temperatura for reduzida de forma muito brusca ao longo das iterações, pode ser que diversas regiões do espaço de busca não sejam exploradas. Por outro lado, se a temperatura for reduzida de forma muito suave, a convergência pode se tornar excessivamente lenta, sendo necessária uma quantidade muito grande de iterações.

---

**Algoritmo 1** Metodologia de Otimização para redes neurais MLP com AAFFFP

---

1: $s_0 \leftarrow$ initial solution
2: $T_0 \leftarrow$ initial temperature
3: Update $s_{best}$ with $s_0$ (best solution found so far)
4: **for** $i = 0$ to $I_{max} - 1$ **do**
5:     **if** $i + 1$ is not a multiple of $I_T$ **then**
6:         $T_{i+1} \leftarrow T_i$
7:     **else**
8:         $T_{i+1} \leftarrow$ new temperature
9:         **if** stopping criteria is satisfied **then**
10:           Stop execution
11:         **end if**
12:     **end if**
13:     Generate a set of $K$ new solutions from $s_i$
14:     Choose the best solution $s'$ from the set
15:     **if** $f(s') < f(s_i)$ **then**
16:         $s_{i+1} \leftarrow s'$
17:     **else**
18:         $s_{i+1} \leftarrow s_0$ with probability $e^{[f(s')-f(s_i)]/T_{i+1}}$
19:     **end if**
20:     Update $s_{best}$ (if $f(s_{i+1}) < f(s_{best})$)
21: Keep the parameter of the AAFFFP contained in $s_{best}$ constant and use the weights and bias as initial ones for training with the backpropagationlearning algorithm with *momentum* and Levenberg-Marquardt learning algorithm.
22: **end for**

---

O método de TS é um algoritmo de busca iterativa caracterizado pelo uso de uma memória flexível [55]. Este método avalia um conjunto de soluções novas a cada iteração (em vez de uma única solução, como acontece em SA) e isto torna TS um método mais rápido, ou seja, necessita de menos iterações do que o SA para convergir. Desta forma, o algoritmo escolhe a nova solução que produz o menor resultado na função de custo, e isto permite que o método escape de mínimos locais. Assim, a melhor solução é sempre aceita como solução atual, ao invés de uma única solução como desempenhado pelo algoritmo SA. O método consiste na geração de uma solução $x_0$ e, em seguida, movimentos aleatórios são gerados na vizinhança desse ponto, com o objetivo de encontrar uma melhor solução para o problema. As soluções geradas são adicionadas à lista tabu, que representa a memória do método, que tem por finalidade impedir a repetição de movimentos

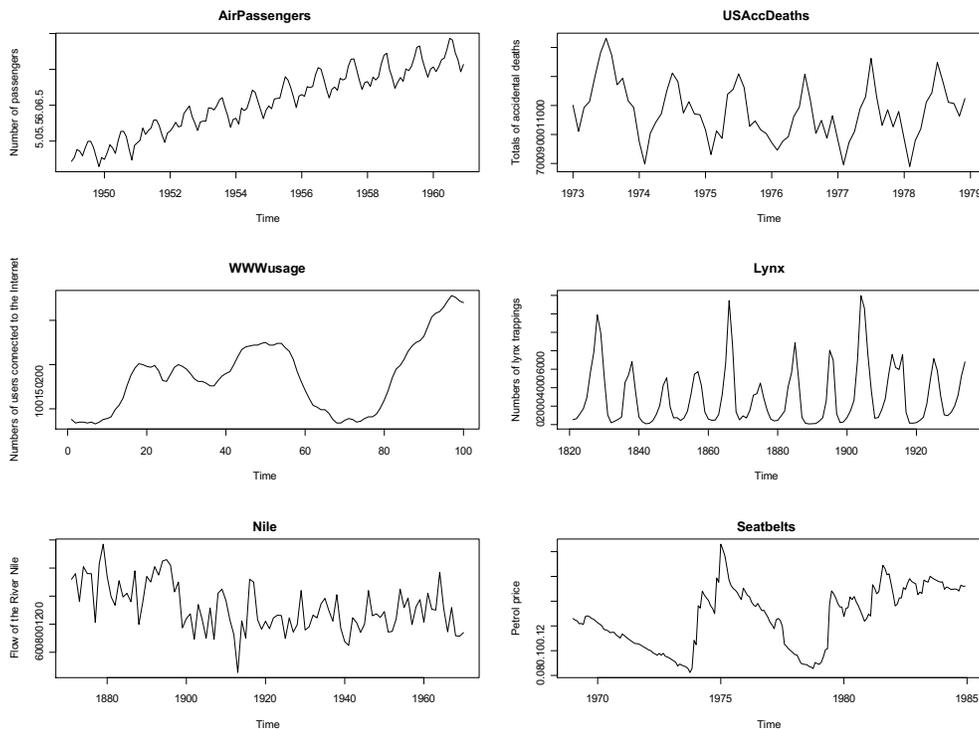

Figura 2. Real time series used.

recentes com o intuito de evitar a geração de soluções repetidas [55].

O pseudo-código da metodologia utilizada está apresentado no Algoritmo 1. A cada iteração, é gerado um conjunto de soluções novas a partir da solução atual, cada uma tem seu custo avaliado e a que apresentar a melhor solução é escolhida, assim como acontece na lista tabu. No entanto, esta solução nem sempre é aceita, diferentemente do que ocorre em TS, pois o critério de aceitação é o mesmo utilizado na técnica de SA. Durante o processo de otimização, armazena-se apenas a melhor solução encontrada e esta é a solução final retornada pelo método. Após encontrar a melhor solução através da combinação da técnicas de SA e TS, mantemos constante o valor do parâmetro $\lambda$, no caso da FFAAPL, contido em $s_{best}$, bem como os valores dos pesos e bias. No caso do uso das outras funções de ativação com parâmetro fixo mantemos constante apenas os valores dos pesos e bias. Esses valores encontrados irão servir de valores iniciais para o treinamento de uma rede MLP com o algoritmo de aprendizagem local. Dois algoritmos foram escolhidos: *backpropagation* com termo *momentum* e o Levenberg-Marquardt. O *backpropagation* com termo *momentum* [4] foi escolhido por ser um dos modelos conexionistas mais utilizados na literatura [32] e o Levenberg-Marquardt foi escolhido por ser um algoritmo projetado para um treinamento rápido sem o uso de uma matriz Hessiana [59]. A descrição original do algoritmo de aprendizagem Levenberg-Marquardt é dado em [60].

Um fator importante é a definição da topologia da rede, em muitas situações, o processo de escolha da arquitetura da rede é feito através de uma sequência de tentativas com diversas topologias. Entretanto, é sabido que se uma topologia tiver uma quantidade pequena de nodos e conexões, a rede pode não ser capaz de representar e aprender os padrões apresentados. Por outro lado, se tiver uma quantidade grande de nodos e conexões, a rede pode conter excesso de parâmetros e apresentar dificuldades para generalização quando forem apresentados padrões ainda não vistos. Portanto, a escolha da topologia de uma rede neural é muito importante, pois influencia fortemente seu desempenho. Por estes motivos é que diversos estudos estão sendo realizados para a otimização de arquitetura da rede. Porém, neste trabalho, para que possamos avaliar os reais efeitos da otimização da função de ativação juntamente com os pesos, optamos por fixar uma topologia com poucos nodos escondidos diminuindo a complexidade da rede.

Neste trabalho, as topologias MLP possuem uma única camada escondida, contendo todas as conexões possíveis entre camadas adjacentes, sem haver conexões entre camadas nãoadjacentes. Portanto, a quantidade de conexões é dada por

$$N = pq + qm$$

onde $p$ é o número de nodos de entrada, $q$ é o número de nodos escondidos e $m$ é o número de nodos de saída.

Considerando um conjunto de soluções $S$ e uma função de custo real $f$, a metodologia utilizada procura o mínimo global $s$, tal que $f(s) \leq f(s'), \forall s' \in S$. A solução inicial $s_0$ é uma rede MLP com uma topologia pré-definida com no máximo 4 nodos escondidos com uso da FFAAPL, o parâmetro $\lambda = 1$, que representa a função logit e os pesos iniciais são extraídos aleatoriamente de uma distribuição uniforme $U(0,1)$. A função de custo é definida por

$$f(s) = \frac{1}{n} \sum_{j=1}^{n} e_j(s) \qquad (7)$$

onde $e_j = t_j - y_j$, $t_j$ e $y_j$ representam, respectivamente, o verdadeiro valor e o valor de saída da rede associados com a $j$-ésima unidade de saída e o padrão de treinamento $i$. O processo termina após $I_{max}$ iterações ou se o critério de parada baseado em validação for satisfeito. Assim, a melhor solução $s_{best}$ encontrada é retornada. O esquema de esfriamento atualiza a temperatura $T_i$ da iteração $i$ a cada $I_T$ iteração do algoritmo. A cada iteração, são geradas $K$ soluções novas a partir da atual. Cada solução contém informações sobre os pesos da rede MLP e, no caso da FFAAPL, o valor do parâmetro $\lambda$.

## V. RESULTADOS EXPERIMENTAIS

Nestes experimentos, usamos uma combinação das técnicas de otimização global SA e TS para otimizar o parâmetro $\lambda$ da família FFAAPL ($F_\lambda$) e os pesos e bias da rede neural. A seguir apresentaremos a descrição dos conjuntos de dados, os parâmetros escolhidos para os experimentos da metodologia de otimização e os resultados encontrados nestes experimentos.

### A. Descrição das bases de dados

Para mostrar a eficácia dos modelos de redes neurais com a família de funções de ativação assimétricas com parâmetro livre, utilizamos seis conjunto de dados de séries temporais com comportamentos não-lineares, ilustradas na Figura 2. Nas séries temporais apresentadas existem características (a priori) importantes para a modelagem, tais como, sazonalidade e tendência constante, assim como séries temporais com comportamentos bastante irregulares, ou seja, séries não-estacionárias, não-sazonais ou sazonais aditiva e multiplicativa, não-Gaussianas e que não apresentam tendência estocástica uniforme. Estes exemplos têm sido usados como benchmarks na literatura de previsão de séries temporais.

*1) Airline passengert:* A primeira série corresponde ao logaritmo do número total de passageiros de uma linha aérea internacional de janeiro de 1949 a dezembro de 1960 (Airline series). A série Airline corresponde aos dados clássicos usados por Box e Jenkins [5] e por Ghiassi, Saidane e Zimbra [61] nos modelos DAN2. A série Airline na sua forma original exibe comportamento não-linear e apresenta comportamento sazonal multiplicativo. Por esta razão, fez-se necessário transformá-los através do logaritmo, com a finalidade de converter a sazonalidade multiplicativa em aditiva. Esta série possui 144 observações e, assim como em diversas pesquisas envolvendo esta série temporal, nós utilizamos os dados dos primeiros 11 anos (132 observações) para ajuste do modelo (conjunto de treinamento) e as últimas 12 observações para previsão (conjunto de teste).

*2) USAccDeaths:* A segunda série corresponde ao número mensal de acidentes com morte nos Estados Unidos no período janeiro de 1973 a dezembro de 1978 (USAccDeaths series). Esses dados foram usados por [62]. A série USAccDeaths exibe um comportamento semelhante à série Airline transformada, porém, não apresenta tendência crescente e não foi necessário fazer transformação nos dados. Esta série possui 72 observações e para o treinamento da rede, utilizamos os primeiros 5 anos (60 observações) e para testar a rede, as últimas 12 observações.

*3) WWWusage:* A terceira série corresponde ao número de usuários conectados na Internet por minuto (WWWusage series) em relação à 100 minutos (100 observações). Na análise desses dados pelos autores Makridakis *et al.* [63], esta série é não-estacionária. Para o treinamento da rede, utilizamos os primeiros 88 primeiros minutos (88 observações) e para testar a rede, as últimas 12 observações.

*4) Lynxt:* A quarta série corresponde ao número de lince canadense preso por ano no distrito de rio de Mackenzie no norte do Canadá para o período 1821-1934 (114 observações). Esta série pode ser obtida em Brockwell e Davis [64] e foi estudada por Campbell e Walker [65] e Zhang [8]. Para o treinamento da rede, utilizamos as primeiras 102 observações e para testar a rede, as últimas 12 observações.

*5) Nile:* A quinta série corresponde às medições da vazão anual do rio Nilo, na Ashwan, no período correspondente a 1871–1970 (100 observações). Esta série foi estudada por Cobb [66] e Balke [67]. Para o treinamento da rede, utilizamos as primeiras 88 medições e para testar a rede, as últimas 12 medições.

*6) PetroPrice:* Finalmente, a sexta e última série corresponde ao preço do petróleo na Grã-Bretanha no período de janeiro de 1969 a dezembro de 1984 (PetroPrice series). Esta série possui 198 observações, sendo que as primeiras 168 observações foram usadas para ajuste dos modelos e as 12 restantes para previsão. Esta série foi usada por Gomes *et al.* [10].

### B. Seleção dos Lags

Para as séries temporais Airline, USAccDeaths, WWWusage, Lynx e PetroPrice, executamos modelos autorregressivos (AR) [5] para selecionar o número de defasagens (lags), essa quantidade selecionada foi usada como nodos de entrada em todos os modelos avaliados. Para a série Nile, o número de defasagem selecionado pelo modelo AR não foi suficiente para o ajuste e previsão do modelo DAN2, por este motivo o número de defasagens escolhido é oito. Os valores selecionados estão apresentados na Tabela I.

Tabela I
RESULT OF SELECTING THE LAGS THROUGH THE AR MODEL.

| Time series | Lags |
|---|---|
| Airline | 5 |
| USAccDeaths | 3 |
| WWWusage | 4 |
| Lynx | 4 |
| Nile | 8 |
| PetroPrice | 3 |

### C. Parâmetros dos Experimentos

Na técnica de otimização global que combina SA e TS (SA+TS), a temperatura inicial é igual a 1 e temperatura é reduzida a cada 10 iterações do algoritmo de otimização de acordo com a equação 6. O número máximo de iterações permitidas é igual a 10,000. Esses valores foram escolhidos empiricamente. Foram realizadas 100 execuções do algoritmo com diferentes inicializações aleatórias de uma uniforme $U(0,1)$ para pesos e bias. O valor de lambda foi inicializado com 1 que representa a função logit. Para cada inicialização, foram realizadas 10 execuções de SA+TS e obtidos os valores médios dessas 10 execuções. O critério de parada $GL_5$ definido em Proben1 [68] também foi utilizado.

O desempenho do algoritmo de SA é influenciado pela escolha do esquema de esfriamento e do mecanismo de geração de soluções novas, entretanto não existem regras objetivas para o ajuste da configuração de modo a obter os melhores resultados possíveis, sendo normalmente adotadas configurações variadas dos parâmetros para avaliar o desempenho [69]. Logo, a configuração adotada neste trabalho foi escolhida empiricamente e pode não ser ótima para o problema abordado. O objetivo desta abordagem é mostrar que o algoritmo de SA alcançou bons resultados para o problema de otimização tratado, apesar da dificuldade para o ajuste dos parâmetros.

Para verificar se o desempenho das redes finais geradas por SA+TS poderia ser melhorado, os valores das conexões e do parâmetro $\lambda$ finais foram usados nas redes MLP e treinadas através dos algoritmos *backpropagation* com o termo *momentum* e Levenberg-Marquardt, que correspondem a`s seguintes codificações SA+TS+BPM e SA+TS+LM, respectivamente. O treinamento é concluído quando atinge 10,000 épocas ou se o erro de validação cresce por 5 épocas consecutivas. A taxa de aprendizagem e o termo *momentum* utilizados foram de 0.001 e 0.9, respectivamente. Uma única camada escondida completamente conectada foi utilizada. Novamente, estes valores podem não ter sido ótimos para o problema, porém, a finalidade deste trabalho é mostrar que é possível melhorar os resultado do ajuste e da previsão de séries temporais das redes geradas por SA+TS com a adição de uma fase de treinamento com um algoritmo de aprendizagem. Para maiores detalhes da arquitetura das redes utilizadas nos diferentes conjunto de dados e do valor médio encontrado para o parâmetro otimizado ($\lambda$) de cada série ver Tabela II.

Como critério de avaliação dos modelos, nós usamos a soma de quadrado dos erros (SSE), o erro quadrático médio (MSE), o erro absoluto médio (MAE) e o erro percentual absoluto médio de previsão (MAPE).

Os resultados e discussões serão apresentados a seguir. Esses resultados serão discutidos de uma forma geral e também em blocos compostos da seguinte forma:

- primeiro bloco: modelos SA+TS(Aranda), SA+TS(Logit) e SA+TS(Cloglog);
- segundo bloco: modelos SA+TS+BPM(Aranda), SA+TS+BPM(Logit) e SA+TS+BPM(Cloglog); e
- terceiro bloco: modelos SA+TS+LM(Aranda), SA+TS+LM(Logit) e SA+TS+LM(Cloglog).

Tabela II

DETALHES DAS ARQUITETURAS UTILIZADAS

| **Airline series** | | | |
|---|---|---|---|
| Activation function | $\lambda$ | Architecture | No. of adjustable parameters |
| Aranda | 2.11 | 5 - 2 - 1 | 16 |
| Logit | 1 | 5 - 2 - 1 | 15 |
| Cloglog | $\to 0$ | 5 - 2 - 1 | 15 |
| **USAccDeaths series** | | | |
| Activation function | $\lambda$ | Architecture | No. of adjustable parameters |
| Aranda | 1.97 | 3 - 3 - 1 | 17 |
| Logit | 1 | 3 - 3 - 1 | 16 |
| Cloglog | $\to 0$ | 3 - 3 - 1 | 16 |
| **WWWusage series** | | | |
| Activation function | $\lambda$ | Architecture | No. of adjustable parameters |
| Aranda | 3.94 | 4 - 4 - 1 | 26 |
| Logit | 1 | 4 - 4 - 1 | 25 |
| Cloglog | $\to 0$ | 4 - 4 - 1 | 25 |
| **Lynx series** | | | |
| Activation function | $\lambda$ | Architecture | No. of adjustable parameters |
| Aranda | 1.76 | 4 - 4 - 1 | 26 |
| Logit | 1 | 4 - 4 - 1 | 25 |
| Cloglog | $\to 0$ | 4 - 4 - 1 | 25 |
| **Nile series** | | | |
| Activation function | $\lambda$ | Architecture | No. of adjustable parameters |
| Aranda | 1.87 | 8 - 4 - 1 | 42 |
| Logit | 1 | 8 - 4 - 1 | 41 |
| Cloglog | $\to 0$ | 8 - 4 - 1 | 41 |
| **PetroPrice series** | | | |
| Activation function | $\lambda$ | Architecture | No. of adjustable parameters |
| Aranda | 1.15 | 3 - 4 - 1 | 22 |
| Logit | 1 | 3 - 4 - 1 | 21 |
| Cloglog | $\to 0$ | 3 - 4 - 1 | 21 |

*D. Resultados e Discussão*

Nas Tabelas III–VIII apresentamos os resultados dos desempenhos médios dos modelos ARIMA, AR e DAN2 e dos desempenhos médios e os respectivos desvios-padrão

Tabela III
RESULTADOS DO DESEMPENHO MÉDIO PARA O AJUSTÉ (CONJUNTO DE TREINAMENTO) E PREVISÃO(CONJUNTO DE TESTE) PARA A SÉRIE AIRLINE.

Train set

| Model | SSE Average | SSE SD | MSE Average | MSE SD | MAE Average | MAE SD | MAPE Average | MAPE SD |
|---|---|---|---|---|---|---|---|---|
| $A$ - ARIMA(4,0,2) | 1.25435 | - | 0.00871 | - | 0.07792 | - | 1.43865 | - |
| $B$ - AR(5) | 1.18442 | - | 0.00933 | - | 0.08479 | - | 1.54486 | - |
| $C$ - DAN2 | 1.12943 | - | 0.00889 | - | 0.08267 | - | 1.50861 | - |
| $D$ - SA+TS (Aranda) | 1.24735 | 2.29E-03 | 0.00982 | 1.80E-05 | 0.08590 | 5.90E-05 | 1.56527 | 1.07E-03 |
| $E$ - SA+TS+BPM (Aranda) | 1.04110 | 2.83E-01 | 0.00820 | 2.23E-03 | 0.07644 | 1.36E-02 | 1.39396 | 2.46E-01 |
| $F$ - SA+TS+LM (Aranda) | 0.83886 | 2.68E-01 | 0.00661 | 2.11E-03 | 0.06649 | 1.32E-02 | 1.22133 | 2.41E-01 |
| $G$ - SA+TS (Logit) | 1.31705 | 1.87E-03 | 0.01037 | 1.47E-05 | 0.08737 | 5.54E-05 | 1.59121 | 1.01E-03 |
| $H$ - SA+TS+BPM (Logit) | 1.15535 | 1.45E-01 | 0.00910 | 1.14E-03 | 0.08057 | 5.69E-03 | 1.46930 | 1.01E-01 |
| $I$ - SA+TS+LM (Logit) | 1.02036 | 2.18E-01 | 0.00803 | 1.72E-03 | 0.07609 | 1.12E-02 | 1.38924 | 2.00E-01 |
| $J$ - SA+TS (Cloglog) | 1.35127 | 3.61E-03 | 0.01064 | 2.84E-05 | 0.08824 | 7.34E-05 | 1.60600 | 1.33E-03 |
| $K$ - SA+TS+BPM (Cloglog) | 1.21830 | 6.66E-02 | 0.00959 | 5.25E-04 | 0.08385 | 1.29E-03 | 1.52769 | 2.42E-02 |
| $L$ - SA+TS+LM (Cloglog) | 0.95010 | 2.05E-01 | 0.00748 | 1.62E-03 | 0.07297 | 1.03E-02 | 1.33071 | 1.87E-01 |

Test set

| Model | SSE Average | SSE SD | MSE Average | MSE SD | MAE Average | MAE SD | MAPE Average | MAPE SD |
|---|---|---|---|---|---|---|---|---|
| $A$ - ARIMA(4,0,2) | 0.31602 | - | 0.02634 | - | 0.14272 | - | 2.31566 | - |
| $B$ - AR(5) | 0.36805 | - | 0.03067 | - | 0.14741 | - | 2.38605 | - |
| $C$ - DAN2 | 0.11481 | - | 0.00957 | - | 0.08392 | - | 1.36551 | - |
| $D$ - SA+TS (Aranda) | 0.18031 | 1.76E-02 | 0.01669 | 1.46E-03 | 0.09628 | 8.99E-04 | 1.87142 | 1.42E-02 |
| $E$ - SA+TS+BPM (Aranda) | 0.14920 | 1.61E-03 | 0.01243 | 1.34E-04 | 0.09085 | 6.80E-04 | 1.46304 | 1.09E-02 |
| $F$ - SA+TS+LM (Aranda) | 0.12777 | 3.78E-03 | 0.01065 | 3.15E-04 | 0.08604 | 1.76E-03 | 1.38931 | 2.83E-02 |
| $G$ - SA+TS (Logit) | 0.16510 | 6.15E-04 | 0.01376 | 5.13E-05 | 0.09726 | 1.60E-04 | 1.56788 | 2.53E-03 |
| $H$ - SA+TS+BPM (Logit) | 0.18628 | 5.54E-03 | 0.01552 | 4.61E-04 | 0.09935 | 1.37E-03 | 1.59407 | 2.17E-02 |
| $I$ - SA+TS+LM (Logit) | 0.13309 | 2.40E-03 | 0.01109 | 2.00E-04 | 0.08915 | 1.05E-03 | 1.43849 | 1.68E-02 |
| $J$ - SA+TS (Cloglog) | 0.16654 | 8.50E-04 | 0.01388 | 7.08E-05 | 0.09764 | 1.74E-04 | 1.57521 | 2.73E-03 |
| $K$ - SA+TS+BPM (Cloglog) | 0.20792 | 3.99E-03 | 0.01733 | 3.32E-04 | 0.10904 | 1.00E-03 | 1.75207 | 1.58E-02 |
| $L$ - SA+TS+LM (Cloglog) | 0.11543 | 4.74E-03 | 0.00962 | 3.95E-04 | 0.08101 | 1.52E-03 | 1.30864 | 2.42E-02 |

referentes as 100 inicializações dos modelos SA+TS, SA+TS+BPM e SA+TS+LM, esses três últimos modelos foram executados com função de ativação Aranda com parâmetro, $\lambda$, livre, com o parâmetro $\lambda = 1$ (que corresponde à função logit) e $\lambda \to 0$ (que corresponde à função complemento log-log). Os modelos ARIMA, AR e DAN2 não apresentam desvios-padrão, pois não tem inicialização aleatória.

Para o modelo ARIMA, experimentamos diferentes modelos do tipo ARIMA($p,1,q$) variando $p = 0,1,2,3,4$ e $q = 0,1,2,3,4$. O melhor modelo de cada série temporal foi selecionado através do menor AIC (Akaike Information Criterion), que é o critério mais comumente utilizado [70].

Nas Tabelas IX, X, XI e XII apresentamos os *p*-valores dos testes *t*-Student[1] para a comparação dos desempenhos médios entre os modelos do segundo bloco e entre os modelos do terceiro bloco, apresentados anteriormente, para as medidas de erro SSE, MSE, MAE e MAPE, respectivamente. Logo, a comparação entre os modelos SA+TS+BPM(Aranda) e SA+TS+BPM(Logit) será representada pela hipótese nula $\mu_E = \mu_H$, entre os modelos SA+TS+BPM(Aranda) e SA+TS+BPM(Cloglog) será representada pela hipótese nula $\mu_E = \mu_K$, entre os modelos SA+TS+BPM(Logit) e SA+TS+BPM(Cloglog) será representada pela hipótese nula $\mu_H = \mu_K$, entre os modelos SA+TS+LM(Aranda) e SA+TS+LM(Logit) será representada pela hipótese nula $\mu_F = \mu_I$, entre os modelos SA+TS+LM(Aranda) e SA+TS+LM(Cloglog) será representada pela hipótese nula $\mu_F = \mu_L$, entre os modelos SA+TS+LM(Logit) e SA+TS+LM(Cloglog) será representada pela hipótese nula $\mu_I = \mu_L$. Para verificar se a diferença entre as médias é estatisticamente significante, o *p*-valor tem que ser menor que o nível de significância, $\alpha$. Neste trabalho, o valor de $\alpha$ é igual a 5% (ou 0.05).

Para a série Airline (Tabela III), podemos observar que, no conjunto de treinamento, o desempenho médio do modelo SA+TS, independentemente da função de ativação usada, foi pior do que o desempenho do modelo DAN2 e ao utilizarmos as funções de ativação logit e complemento log-log seu desempenho médio foi pior do que o desempenho do modelo ARIMA. Com o uso do algoritmo de aprendizagem local BPM para fazer um ajuste local, apenas o desempenho do modelo SA+TS+BPM(Aranda) superou o desempenho do modelo DAN2 e com o uso do algoritmo de aprendizagem local LM, o desempenho do modelo SA+TS+BPM, independentemente da função de ativação usada, foi melhor do que o modelo DAN2. O ajuste apresentado pelo modelo SA+TS+BPM(Aranda) foi melhor do que todos os outros modelos. No conjunto de teste, podemos notar que o desempenho melhora com o uso dos modelos SA+TS+BPM e SA+TS+LM, independentemente da função de ativação, sendo que o modelo SA+TS+LM(Aranda) apresentou o melhor resultado em relação aos outros modelos, exceto em relação ao modelo DAN2 em que os resultados foram equivalentes.

---

[1] O teste *t*-Student é um teste paramétrico utilizado na estatística para comparar duas ou mais amostras independentes com a finalidade de verificar a existência de diferença significativa entre as médias de métricas dessas amostras [71].

Para a série USAccDeaths (Tabela IV), nos conjuntos de treinamento e de teste, todos os modelos SA+TS, SA+TS+BPM e SA+TS+LM apresentaram desempenhos médios superiores em relação aos desempenhos dos modelos ARIMA, AR e DAN2. Podemos observar que os modelos que combinam as técnicas de otimização global e otimização local apresentam melhorias em seus resultados médios em relação aos modelos que usam apenas técnicas de otimização global. O modelo SA+TS+LM(Aranda) apresentou o melhor resultado.

Para a série WWWusage (Tabela V), podemos observar que os modelos que combinam as técnicas de otimização global e otimização local apresentam melhorias em seus resultados médios em relação aos modelos que usam apenas técnicas de otimização global. No conjunto de treinamento, apenas os modelos SA+TS+LM conseguem resultados melhores do que o modelo AR, o qual obteve desempenho superior ao apresentado pelo modelo DAN2. No conjunto de teste, o modelo DAN2 superou quase todos os modelos em questão, exceto o modelo SA+TS+LM(Aranda). Portanto, o desempenho do modelo SA+TS+LM(Aranda) foi superior aos desempenhos de todos os outros modelos.

Para a série Lynx (Tabela VI), os modelos SA+TS+BPM e SA+TS+LM apresentaram melhorias substanciais em seus resultados médios em relação aos modelos que usam apenas técnicas de otimização global (SA+TS). No conjunto de treinamento, todos os modelos SA+TS, SA+TS+BPM e SA+TS+LM apresentaram melhores desempenhos médios do que os desempenhos dos modelos ARIMA, AR e DAN2. No conjunto de teste, apenas os modelos com função de ativação Aranda apresentaram desempenhos superiores aos modelos ARIMA, AR e DAN2. O modelo SA+TS+LM(Aranda) apresentou o melhor resultado em relação a todos os outros modelos.

Para a série Nile (Tabela VII), apenas os modelos SA+TS+LM, independentemente da função de ativação, apresentaram melhores desempenhos médios em relação ao modelo DAN2 no conjunto de treinamento. No conjunto de teste, apenas os modelos SA+TS+LM(Aranda) e SA+TS+LM(Cloglog) apresentaram desempenhos médios superiores ao apresentado pelo modelo DAN2. Vale ressaltar que os modelos que combinam as técnicas de otimização global e otimização local apresentam melhorias em seus resultados médios em relação aos modelos que usam apenas técnicas de otimização global e que o desempenho do modelo SA+TS+LM(Aranda) foi superior aos desempenhos de todos os outros modelos, tanto no conjunto de treinamento quanto no conjunto de teste.

Finalmente, para a série PetroPrice (Tabela VIII), nos conjuntos de treinamento e de teste, observamos que os modelos SA+TS+BPM e SA+TS+LM apresentaram melhorias substanciais em seus desempenhos médios em relação aos modelos SA+TS e que todos os desempenhos médios obtidos pelos modelos com a metodologia utilizada foram superiores aos desempenhos obtidos pelos modelos ARIMA, AR e DAN2. O desempenho do modelo SA+TS+LM(Aranda) foi superior aos desempenhos de todos os outros modelos.

## VI. Conclusões

Os resultados apresentados mostram que os modelos que combinam as técnicas de otimização global e otimização local apresentam melhorias em seus resultados médios em relação aos modelos que usam apenas técnicas de otimização global, mostram ainda que o desempenho do modelo SA+TS+LM(Aranda). Nos seis exemplos de séries temporais estudados, foi superior aos desempenhos de todos os outros modelos, inclusive em relação aos modelos DAN2, que são modelos extremamente eficientes no ajuste e previsão de séries temporais.

Para todos os problemas abordados, ao compararmos os modelos por blocos, observamos que os modelos com função de ativação Aranda apresentam desempenhos médios melhores do que os modelos com função de ativação logit e complemento log-log e essa diferença é estatisticamente significante uma vez que todos os $p$-valores apresentados nas Tabelas IX–XII são menores que 0.05.

Portanto, podemos concluir que a implementação de uma metodologia combinando as principais características favoráveis dos algoritmos de SA e TS, fazendo uso de um algoritmo local de aprendizagem, é capaz de produzir resultados bastante satisfatórios para otimização da função de ativação e pesos de redes MLP para os problemas de séries temporais abordados. Vale ressaltar que todos os resultados apresentados podem não ter sido ótimos para cada problema, ou seja, pode ser que os modelos ARIMA, AR e DAN2 alcancem resultados melhores do que os que foram apresentados neste estudo apenas alterando o número de lags, porém, o objetivo desta abordagem é mostrar que é possível melhorar os resultados de ajuste e previsão de séries temporais das redes geradas por SA e TS com a introdução de uma fase de treinamento com o algoritmo de aprendizagem local, *backpropagation* com *momentum* ou Levenberg-Marquardt, apesar das dificuldades no ajuste dos parâmetros dos algoritmos.

Tabela IV
RESULTADOS DO DESEMPENHO MÉDIO PARA O AJUSTE (CONJUNTO DE TREINAMENTO) E PREVISÃO (CONJUNTO DE TESTE) PARA A SÉRIE USACCDEATHS.

| Model | Train set | | | | | | | |
|---|---|---|---|---|---|---|---|---|
| | SSE | | MSE | | MAE | | MAPE | |
| | Average | SD | Average | SD | Average | SD | Average | SD |
| *A* - ARIMA(4,0,3) | 23524333 | - | 326727 | - | 477 | - | 5.32 | - |
| *B* - AR(3) | 23067905 | - | 404700 | - | 527 | - | 6.09 | - |
| *C* - DAN2 | 22621878 | - | 396875 | - | 509 | - | 5.93 | - |
| *D* - SA+TS (Aranda) | 17821120 | 2.86E+04 | 312651 | 5.02E+02 | 452 | 3.89E-01 | 5.18 | 4.54E-03 |
| *E* - SA+TS+BPM (Aranda) | 16046939 | 2.54E+06 | 281525 | 4.46E+04 | 424 | 3.88E+01 | 4.90 | 4.49E-01 |
| *F* - SA+TS+LM (Aranda) | 12260440 | 4.41E+06 | 215095 | 7.74E+04 | 355 | 7.53E+01 | 4.13 | 8.44E-01 |
| *G* - SA+TS (Logit) | 19793697 | 2.35E+04 | 347258 | 4.12E+02 | 484 | 4.05E-01 | 5.55 | 4.71E-03 |
| *H* - SA+TS+BPM (Logit) | 18010923 | 3.99E+06 | 315981 | 7.01E+04 | 457 | 6.57E+01 | 5.27 | 7.44E-01 |
| *I* - SA+TS+LM (Logit) | 15425933 | 2.32E+06 | 270630 | 4.06E+04 | 421 | 3.33E+01 | 4.84 | 3.63E-01 |
| *J* - SA+TS (Cloglog) | 20665241 | 8.78E+03 | 362548 | 1.54E+02 | 490 | 2.07E-01 | 5.64 | 2.41E-03 |
| *K* - SA+TS+BPM (Cloglog) | 19639433 | 3.48E+06 | 344551 | 6.10E+04 | 476 | 4.99E+01 | 5.49 | 5.87E-01 |
| *L* - SA+TS+LM (Cloglog) | 18411535 | 4.63E+06 | 323009 | 8.12E+04 | 454 | 6.39E+01 | 5.20 | 7.07E-01 |

| Model | Test set | | | | | | | |
|---|---|---|---|---|---|---|---|---|
| | SSE | | MSE | | MAE | | MAPE | |
| | Average | SD | Average | SD | Average | SD | Average | SD |
| *A* - ARIMA(4,0,3) | 11308630 | - | 942386 | - | 803 | - | 9.47 | - |
| *B* - AR(3) | 13315781 | - | 1109648 | - | 861 | - | 10.22 | - |
| *C* - DAN2 | 4880680 | - | 406723 | - | 530 | - | 6.25 | - |
| *D* - SA+TS (Aranda) | 3623262 | 5.88E+03 | 301938 | 4.90E+02 | 429 | 4.01E-01 | 5.06 | 4.87E-03 |
| *E* - SA+TS+BPM (Aranda) | 3276890 | 8.61E+05 | 273074 | 7.17E+04 | 402 | 7.11E+01 | 4.70 | 8.66E-01 |
| *F* - SA+TS+LM (Aranda) | 3016732 | 1.17E+06 | 251394 | 9.78E+04 | 386 | 8.10E+01 | 4.55 | 1.03E+00 |
| *G* - SA+TS (Logit) | 3853019 | 7.52E+03 | 321085 | 6.27E+02 | 455 | 3.72E-01 | 5.35 | 4.28E-03 |
| *H* - SA+TS+BPM (Logit) | 3769943 | 3.59E+05 | 314162 | 2.99E+04 | 438 | 3.73E+01 | 5.19 | 4.90E-01 |
| *I* - SA+TS+LM (Logit) | 3699661 | 5.12E+05 | 308305 | 4.27E+04 | 428 | 3.37E+01 | 4.99 | 4.27E-01 |
| *J* - SA+TS (Cloglog) | 4528474 | 3.10E+03 | 377373 | 2.58E+02 | 483 | 2.87E-01 | 5.75 | 2.97E-03 |
| *K* - SA+TS+BPM (Cloglog) | 4026091 | 5.89E+05 | 335508 | 4.91E+04 | 453 | 3.92E+01 | 5.34 | 4.97E-01 |
| *L* - SA+TS+LM (Cloglog) | 4010641 | 7.97E+05 | 324220 | 6.64E+04 | 451 | 4.61E+01 | 5.26 | 5.63E-01 |

SD - Standard deviation.

Tabela V
Resultados do desempenho médio para o ajuste (conjunto de treinamento) e previsão(conjunto de teste) para a série WWWusage.

| Model | Train set | | | | | | | |
|---|---|---|---|---|---|---|---|---|
| | SSE | | MSE | | MAE | | MAPE | |
| | Average | SD | Average | SD | Average | SD | Average | SD |
| *A* - ARIMA(4,0,1) | 926.60 | - | 9.27 | - | 2.37 | - | 2.06 | - |
| *B* - AR(4) | 728.34 | - | 8.67 | - | 2.31 | - | 1.91 | - |
| *C* - DAN2 | 731.17 | - | 8.70 | - | 2.31 | - | 1.90 | - |
| *D* - SA+TS (Aranda) | 987.50 | 1.40E+00 | 11.76 | 1.67E-02 | 2.80 | 1.78E-03 | 2.24 | 1.29E-03 |
| *E* - SA+TS+BPM (Aranda) | 765.32 | 8.81E+01 | 9.11 | 1.05E+00 | 2.41 | 1.83E-01 | 1.97 | 1.38E-01 |
| *F* - SA+TS+LM (Aranda) | 669.46 | 1.97E+01 | 7.97 | 2.35E-01 | 2.24 | 3.14E-02 | 1.85 | 2.17E-02 |
| *G* - SA+TS (Logit) | 1202.87 | 3.22E+00 | 14.32 | 3.84E-02 | 3.04 | 3.47E-03 | 2.44 | 3.02E-03 |
| *H* - SA+TS+BPM (Logit) | 986.53 | 4.51E+02 | 11.74 | 5.37E+00 | 2.70 | 5.80E-01 | 2.16 | 4.36E-01 |
| *I* - SA+TS+LM (Logit) | 690.50 | 3.51E+01 | 8.22 | 4.18E-01 | 2.27 | 5.96E-02 | 1.86 | 4.70E-02 |
| *J* - SA+TS (Cloglog) | 1320.32 | 6.69E+00 | 15.72 | 7.97E-02 | 3.14 | 6.73E-03 | 2.57 | 5.93E-03 |
| *K* - SA+TS+BPM (Cloglog) | 989.44 | 3.58E+02 | 11.78 | 4.26E+00 | 2.73 | 4.10E-01 | 2.20 | 2.83E-01 |
| *L* - SA+TS+LM (Cloglog) | 722.07 | 8.94E+01 | 8.60 | 1.06E+00 | 2.34 | 1.63E-01 | 1.91 | 1.21E-01 |

| Model | Test set | | | | | | | |
|---|---|---|---|---|---|---|---|---|
| | SSE | | MSE | | MAE | | MAPE | |
| | Average | SD | Average | SD | Average | SD | Average | SD |
| *A* - ARIMA(4,0,1) | 4507.32 | - | 375.61 | - | 14.74 | - | 7.55 | - |
| *B* - AR(4) | 43939.64 | - | 3661.64 | - | 52.21 | - | 24.31 | - |
| *C* - DAN2 | 149.11 | - | 12.43 | - | 3.12 | - | 1.51 | - |
| *D* - SA+TS (Aranda) | 1309.51 | 1.04E+01 | 109.13 | 8.65E-01 | 8.76 | 3.51E-02 | 4.11 | 1.62E-02 |
| *E* - SA+TS+BPM (Aranda) | 302.06 | 1.07E+02 | 25.17 | 8.88E+00 | 4.28 | 3.63E-01 | 2.05 | 1.65E-01 |
| *F* - SA+TS+LM (Aranda) | 144.77 | 3.23E+01 | 12.06 | 2.69E+00 | 3.05 | 1.98E-01 | 1.47 | 9.07E-02 |
| *G* - SA+TS (Logit) | 2747.52 | 3.09E+01 | 228.96 | 2.57E+00 | 11.59 | 6.73E-02 | 5.40 | 3.06E-02 |
| *H* - SA+TS+BPM (Logit) | 604.44 | 4.94E+02 | 50.37 | 4.12E+01 | 6.34 | 6.77E-01 | 3.00 | 3.14E-01 |
| *I* - SA+TS+LM (Logit) | 206.99 | 1.59E+02 | 17.25 | 1.33E+01 | 3.18 | 4.52E-01 | 1.53 | 2.04E-01 |
| *J* - SA+TS (Cloglog) | 753.38 | 7.15E+00 | 62.78 | 5.96E-01 | 6.47 | 3.12E-02 | 3.07 | 1.44E-02 |
| *K* - SA+TS+BPM (Cloglog) | 165.96 | 3.15E+02 | 13.83 | 2.63E+01 | 3.01 | 7.52E-01 | 1.46 | 3.45E-01 |
| *L* - SA+TS+LM (Cloglog) | 196.42 | 1.57E+02 | 16.37 | 1.31E+01 | 3.18 | 5.39E-01 | 1.52 | 2.46E-01 |

SD - Standard deviation.

Tabela VI
RESULTADOS DO DESEMPENHO MÉDIO PARA O AJUSTE (CONJUNTO DE TREINAMENTO) E PREVISÃO (CONJUNTO DE TESTE) PARA A SÉRIE LYNX.

| Model | Train set | | | | | | | |
|---|---|---|---|---|---|---|---|---|
| | SSE | | MSE | | MAE | | MAPE | |
| | Average | SD | Average | SD | Average | SD | Average | SD |
| $A$ - ARIMA(4,0,4) | 81774703 | - | 717322 | - | 604 | - | 135.52 | - |
| $B$ - AR(4) | 78817882 | - | 804264 | - | 635 | - | 140.91 | - |
| $C$ - DAN2 | 78506301 | - | 801085 | - | 633 | - | 154.22 | - |
| $D$ - SA+TS (Aranda) | 67979113 | 7.66E+05 | 693664 | 7.82E+03 | 587 | 6.23E+00 | 110.65 | 1.14E+00 |
| $E$ - SA+TS+BPM (Aranda) | 53317504 | 1.64E+05 | 544056 | 1.67E+03 | 495 | 1.17E+00 | 66.81 | 5.07E-01 |
| $F$ - SA+TS+LM (Aranda) | 44699512 | 1.19E+05 | 456117 | 1.21E+03 | 453 | 7.24E-01 | 58.42 | 3.50E-01 |
| $G$ - SA+TS (Logit) | 69057879 | 7.75E+05 | 704672 | 7.91E+03 | 591 | 6.21E+00 | 111.58 | 1.22E+00 |
| $H$ - SA+TS+BPM (Logit) | 56428083 | 2.30E+05 | 575797 | 2.35E+03 | 582 | 1.22E+00 | 67.81 | 4.76E-01 |
| $I$ - SA+TS+LM (Logit) | 59326087 | 2.22E+05 | 605368 | 2.26E+03 | 507 | 1.07E+00 | 71.13 | 4.18E-01 |
| $J$ - SA+TS (Cloglog) | 72179787 | 7.81E+05 | 736528 | 7.97E+03 | 600 | 6.27E+00 | 115.67 | 1.21E+00 |
| $K$ - SA+TS+BPM (Cloglog) | 59983420 | 2.19E+05 | 612076 | 2.24E+03 | 505 | 1.34E+00 | 70.63 | 5.83E-01 |
| $L$ - SA+TS+LM (Cloglog) | 51160480 | 1.88E+05 | 520005 | 1.92E+03 | 477 | 9.81E-01 | 62.73 | 3.44E-01 |

| Model | Test set | | | | | | | |
|---|---|---|---|---|---|---|---|---|
| | SSE | | MSE | | MAE | | MAPE | |
| | Average | SD | Average | SD | Average | SD | Average | SD |
| $A$ - ARIMA(4,0,4) | 19573995 | - | 1631166 | - | 1093 | - | 77.24 | - |
| $B$ - AR(4) | 19661535 | - | 1638461 | - | 1101 | - | 77.95 | - |
| $C$ - DAN2 | 1756710 | - | 146393 | - | 325 | - | 21.77 | - |
| $D$ - SA+TS (Aranda) | 1422224 | 3.19E+04 | 118519 | 2.66E+03 | 262 | 3.83E+00 | 18.35 | 3.01E-01 |
| $E$ - SA+TS+BPM (Aranda) | 1388380 | 1.59E+04 | 112365 | 1.33E+03 | 260 | 7.34E-01 | 17.48 | 9.02E-02 |
| $F$ - SA+TS+LM (Aranda) | 712973 | 1.06E+04 | 59414 | 8.87E+02 | 204 | 6.29E-01 | 15.09 | 2.91E-02 |
| $G$ - SA+TS (Logit) | 2107432 | 2.02E+04 | 175619 | 1.68E+03 | 324 | 3.08E+00 | 22.67 | 2.29E-01 |
| $H$ - SA+TS+BPM (Logit) | 1978871 | 1.30E+04 | 161573 | 1.08E+03 | 279 | 1.12E+00 | 18.83 | 6.73E-02 |
| $I$ - SA+TS+LM (Logit) | 1586155 | 2.36E+04 | 148846 | 1.96E+03 | 265 | 1.62E+00 | 17.20 | 9.49E-02 |
| $J$ - SA+TS (Cloglog) | 1831049 | 1.96E+04 | 172587 | 1.64E+03 | 294 | 3.05E+00 | 20.01 | 2.27E-01 |
| $K$ - SA+TS+BPM (Cloglog) | 1786869 | 1.47E+04 | 165572 | 1.22E+03 | 291 | 1.17E+00 | 18.02 | 9.42E-02 |
| $L$ - SA+TS+LM (Cloglog) | 1386327 | 1.46E+04 | 115527 | 1.22E+03 | 260 | 4.18E-01 | 17.05 | 5.43E-02 |

SD - Standard deviation.

Tabela VII

RESULTADOS DO DESEMPENHO MÉDIO PARA O AJUSTE (CONJUNTO DE TREINAMENTO) E PREVISÃO (CONJUNTO DE TESTE) PARA A SÉRIE NILE.

| Model | Train set | | | | | | | |
|---|---|---|---|---|---|---|---|---|
| | SSE | | MSE | | MAE | | MAPE | |
| | Average | SD | Average | SD | Average | SD | Average | SD |
| *A* - ARIMA(3,0,2) | 1706634 | - | 17066 | - | 108 | - | 12.04 | - |
| *B* - AR(4) | 1760539 | - | 20471 | - | 116 | - | 13.19 | - |
| *C* - DAN2 | 1422972 | - | 17787 | - | 110 | - | 12.61 | - |
| *D* - SA+TS (Aranda) | 1679142 | 3.74E+02 | 19525 | 4.35E+00 | 112 | 2.05E-02 | 12.80 | 2.30E-03 |
| *E* - SA+TS+BPM (Aranda) | 1452418 | 2.15E+04 | 18155 | 2.69E+02 | 108 | 6.76E-01 | 12.49 | 6.15E-02 |
| *F* - SA+TS+LM (Aranda) | 947082 | 3.09E+04 | 11839 | 3.86E+02 | 85 | 1.33E+00 | 10.10 | 1.40E-01 |
| *G* - SA+TS (Logit) | 1719719 | 4.48E+02 | 19997 | 5.20E+00 | 113 | 3.15E-02 | 12.84 | 3.35E-03 |
| *H* - SA+TS+BPM (Logit) | 1573466 | 3.79E+04 | 19668 | 4.74E+02 | 110 | 1.00E+00 | 12.57 | 7.17E-02 |
| *I* - SA+TS+LM (Logit) | 1025089 | 5.21E+04 | 12814 | 6.51E+02 | 91 | 1.80E+00 | 10.65 | 1.68E-01 |
| *J* - SA+TS (Cloglog) | 1681252 | 6.79E+02 | 19549 | 7.89E+00 | 111 | 3.30E-02 | 12.67 | 3.41E-03 |
| *K* - SA+TS+BPM (Cloglog) | 1635122 | 1.53E+04 | 20439 | 1.92E+02 | 114 | 5.06E-01 | 12.99 | 6.66E-02 |
| *L* - SA+TS+LM (Cloglog) | 1283143 | 2.49E+04 | 16039 | 3.11E+02 | 104 | 6.29E-01 | 12.00 | 5.74E-02 |
| Model | Test set | | | | | | | |
| | SSE | | MSE | | MAE | | MAPE | |
| | Average | SD | Average | SD | Average | SD | Average | SD |
| *A* - ARIMA(3,0,2) | 311370 | - | 25948 | - | 129 | - | 13.58 | - |
| *B* - AR(4) | 262563 | - | 21880 | - | 118 | - | 13.77 | - |
| *C* - DAN2 | 175212 | - | 14601 | - | 104 | - | 12.10 | - |
| *D* - SA+TS (Aranda) | 237107 | 1.55E+02 | 19759 | 1.29E+01 | 118 | 3.76E-02 | 13.53 | 4.46E-03 |
| *E* - SA+TS+BPM (Aranda) | 204063 | 6.96E+02 | 17005 | 5.80E+01 | 109 | 3.75E-01 | 12.85 | 4.24E-02 |
| *F* - SA+TS+LM (Aranda) | 160088 | 1.96E+03 | 13062 | 1.64E+02 | 100 | 7.42E-01 | 11.75 | 9.41E-02 |
| *G* - SA+TS (Logit) | 226725 | 6.86E+01 | 18894 | 5.71E+00 | 118 | 8.28E-03 | 13.53 | 6.78E-04 |
| *H* - SA+TS+BPM (Logit) | 228603 | 9.20E+02 | 19050 | 7.67E+01 | 113 | 3.04E-01 | 13.12 | 5.15E-02 |
| *I* - SA+TS+LM (Logit) | 183932 | 2.86E+03 | 16994 | 2.38E+02 | 107 | 4.58E-01 | 12.43 | 6.80E-02 |
| *J* - SA+TS (Cloglog) | 229351 | 7.03E+01 | 19113 | 5.86E+00 | 118 | 1.23E-02 | 13.56 | 1.46E-03 |
| *K* - SA+TS+BPM (Cloglog) | 208317 | 4.90E+03 | 17360 | 4.09E+02 | 111 | 1.34E+00 | 12.98 | 1.70E-01 |
| *L* - SA+TS+LM (Cloglog) | 160748 | 3.02E+03 | 13341 | 2.52E+02 | 102 | 9.22E-01 | 11.83 | 1.03E-01 |

SD - Standard deviation.

Tabela VIII
RESULTADOS DO DESEMPENHO MÉDIO PARA O AJUSTE (CONJUNTO DE TREINAMENTO) E PREVISÃO (CONJUNTO DE TESTE) PARA A SÉRIE PETROPRICE.

| Model | Train set | | | | | | | |
|---|---|---|---|---|---|---|---|---|
| | SSE* | | MSE* | | MAE* | | MAPE | |
| | Average | SD | Average | SD | Average | SD | Average | SD |
| $A$ - ARIMA(4,0,3) | 16.64274 | - | 0.08668 | - | 17.7231 | - | 1.66 | - |
| $B$ - AR(3) | 17.40503 | - | 0.09833 | - | 19.0923 | - | 1.81 | - |
| $C$ - DAN2 | 17.14210 | - | 0.00097 | - | 0.19164 | - | 1.82 | - |
| $D$ - SA+TS (Aranda) | 16.64342 | 2.16E-03 | 0.00094 | 1.22E-07 | 0.18691 | 2.20E-05 | 1.77 | 2.10E-04 |
| $E$ - SA+TS+BPM (Aranda) | 14.82348 | 1.67E-02 | 0.00084 | 9.45E-07 | 0.18035 | 1.65E-04 | 1.71 | 1.61E-03 |
| $F$ - SA+TS+LM (Aranda) | 11.99321 | 1.67E-02 | 0.00068 | 9.42E-07 | 0.17088 | 5.84E-05 | 1.62 | 5.78E-04 |
| $G$ - SA+TS (Logit) | 16.90719 | 5.65E-03 | 0.00096 | 3.19E-07 | 0.19023 | 1.03E-04 | 1.80 | 1.06E-03 |
| $H$ - SA+TS+BPM (Logit) | 15.82313 | 3.16E-02 | 0.00089 | 1.78E-06 | 0.18069 | 3.01E-04 | 1.71 | 3.00E-03 |
| $I$ - SA+TS+LM (Logit) | 15.44169 | 3.06E-03 | 0.00087 | 1.73E-07 | 0.17250 | 5.52E-05 | 1.63 | 5.74E-04 |
| $J$ - SA+TS (Cloglog) | 16.78573 | 1.11E-02 | 0.00095 | 6.25E-07 | 0.18787 | 8.84E-05 | 1.78 | 8.16E-04 |
| $K$ - SA+TS+BPM (Cloglog) | 15.82730 | 2.72E-02 | 0.00091 | 1.54E-06 | 0.18545 | 2.27E-04 | 1.69 | 2.27E-03 |
| $L$ - SA+TS+LM (Cloglog) | 14.91973 | 3.90E-02 | 0.00084 | 2.20E-06 | 0.18619 | 4.87E-04 | 1.76 | 5.46E-03 |

| Model | Test set | | | | | | | |
|---|---|---|---|---|---|---|---|---|
| | SSE* | | MSE* | | MAE* | | MAPE | |
| | Average | SD | Average | SD | Average | SD | Average | SD |
| $A$ - ARIMA(4,0,3) | 0.09530 | - | 0.00794 | - | 7.47363 | - | 0.65 | - |
| $B$ - AR(3) | 2.34080 | - | 0.19507 | - | 37.9905 | - | 3.29 | - |
| $C$ - DAN2 | 0.00188 | - | 0.00016 | - | 0.09067 | - | 0.78 | - |
| $D$ - SA+TS (Aranda) | 0.00162 | 1.10E-06 | 0.00013 | 9.20E-08 | 0.08649 | 4.33E-05 | 0.75 | 3.73E-04 |
| $E$ - SA+TS+BPM (Aranda) | 0.00152 | 1.19E-06 | 0.00013 | 9.90E-08 | 0.08064 | 1.03E-04 | 0.70 | 8.97E-04 |
| $F$ - SA+TS+LM (Aranda) | 0.00148 | 3.59E-07 | 0.00012 | 3.00E-08 | 0.07930 | 1.07E-05 | 0.69 | 9.18E-05 |
| $G$ - SA+TS (Logit) | 0.00168 | 1.14E-06 | 0.00014 | 9.50E-08 | 0.09136 | 3.90E-05 | 0.79 | 3.41E-04 |
| $H$ - SA+TS+BPM (Logit) | 0.00163 | 2.01E-06 | 0.00014 | 1.67E-07 | 0.08388 | 1.05E-04 | 0.73 | 9.14E-04 |
| $I$ - SA+TS+LM (Logit) | 0.00151 | 1.53E-06 | 0.00013 | 1.27E-07 | 0.07519 | 6.46E-05 | 0.65 | 5.55E-04 |
| $J$ - SA+TS (Cloglog) | 0.00160 | 2.67E-06 | 0.00013 | 2.22E-07 | 0.08353 | 9.55E-05 | 0.72 | 8.16E-04 |
| $K$ - SA+TS+BPM (Cloglog) | 0.00150 | 3.58E-06 | 0.00013 | 2.99E-07 | 0.08236 | 1.06E-04 | 0.72 | 9.27E-04 |
| $L$ - SA+TS+LM (Cloglog) | 0.00153 | 7.36E-07 | 0.00013 | 6.10E-08 | 0.08037 | 3.90E-05 | 0.70 | 3.43E-04 |

SD - Standard deviation.

*Values multiplied by $10^4$ for better visualization.

Tabela IX
RESULTADOS DOS $p$-VALORES DOS TESTES $t$-STUDENT COM NÍVEL DE 5% DE SIGNIFICANCIA PARA A MEDIDA DE ERRO SSE.

| Time series | Comparison training set | | | | | |
|---|---|---|---|---|---|---|
| | $\mu E = \mu H$ | $\mu E = \mu K$ | $\mu H = \mu K$ | $\mu F = \mu I$ | $\mu F = \mu L$ | $\mu I = \mu L$ |
| Airline | 0.0004 | 0.0000 | 0.0001 | 0.0000 | 0.0012 | 0.0200 |
| USAccDeaths | 0.0000 | 0.0000 | 0.0024 | 0.0000 | 0.0000 | 0.0000 |
| WWWusage | 0.0000 | 0.0000 | 0.9597 | 0.0000 | 0.0000 | 0.0012 |
| Lynx | 0.0000 | 0.0000 | 0.0000 | 0.0000 | 0.0000 | 0.0000 |
| Nile | 0.0000 | 0.0000 | 0.0000 | 0.0000 | 0.0000 | 0.0000 |
| PetroPrice | 0.0000 | 0.0000 | 0.3180 | 0.0000 | 0.0000 | 0.0000 |
| Time series | Comparison test set | | | | | |
| | $\mu E = \mu H$ | $\mu E = \mu K$ | $\mu H = \mu K$ | $\mu F = \mu I$ | $\mu F = \mu L$ | $\mu I = \mu L$ |
| Airline | 0.0000 | 0.0000 | 0.0000 | 0.0000 | 0.0000 | 0.0000 |
| USAccDeaths | 0.0000 | 0.0000 | 0.0003 | 0.0000 | 0.0000 | 0.0012 |
| WWWusage | 0.0000 | 0.0001 | 0.0000 | 0.0002 | 0.0015 | 0.6374 |
| Lynx | 0.0000 | 0.0000 | 0.0000 | 0.0000 | 0.0000 | 0.0000 |
| Nile | 0.0000 | 0.0000 | 0.0000 | 0.0000 | 0.0686 | 0.0000 |
| PetroPrice | 0.0000 | 0.0000 | 0.0000 | 0.0000 | 0.0000 | 0.0000 |

Tabela X
RESULTADOS DOS $p$-VALORES DOS TESTES $t$-STUDENT COM NÍVEL DE 5% DE SIGNIFICANCIA PARA A MEDIDA DE ERRO MSE.

| Time series | Comparison training set | | | | | |
|---|---|---|---|---|---|---|
| | $\mu E = \mu H$ | $\mu E = \mu K$ | $\mu H = \mu K$ | $\mu F = \mu I$ | $\mu F = \mu L$ | $\mu I = \mu L$ |
| Airline | 0.0000 | 0.0002 | 0.0000 | 0.0011 | 0.0000 | 0.0369 |
| USAccDeaths | 0.0000 | 0.0000 | 0.0024 | 0.0000 | 0.0000 | 0.0000 |
| WWWusage | 0.0000 | 0.0000 | 0.9597 | 0.0000 | 0.0000 | 0.0012 |
| Lynx | 0.0000 | 0.0000 | 0.0000 | 0.0000 | 0.0000 | 0.0000 |
| Nile | 0.0000 | 0.0000 | 0.8507 | 0.0000 | 0.0000 | 0.0000 |
| PetroPrice | 0.0000 | 0.0000 | 0.0000 | 0.0000 | 0.0000 | 0.0000 |
| Time series | Comparison test set | | | | | |
| | $\mu E = \mu H$ | $\mu E = \mu K$ | $\mu H = \mu K$ | $\mu F = \mu I$ | $\mu F = \mu L$ | $\mu I = \mu L$ |
| Airline | 0.0000 | 0.0000 | 0.0000 | 0.0000 | 0.0000 | 0.0000 |
| USAccDeaths | 0.0000 | 0.0000 | 0.0003 | 0.0000 | 0.0000 | 0.0452 |
| WWWusage | 0.0000 | 0.0001 | 0.0000 | 0.0002 | 0.0015 | 0.6374 |
| Lynx | 0.0000 | 0.0000 | 0.0000 | 0.0000 | 0.0000 | 0.0000 |
| Nile | 0.0000 | 0.0000 | 0.0000 | 0.0000 | 0.0000 | 0.0000 |
| PetroPrice | 0.0000 | 0.0000 | 0.0000 | 0.0000 | 0.0000 | 0.0000 |

Tabela XI

RESULTADOS DOS $p$-VALORES DOS TESTES $t$-STUDENT COM NÍVEL DE 5% DE SIGNIFICANCIA PARA A MEDIDA DE ERRO MAE.

| Time series | Comparison training set | | | | | |
|---|---|---|---|---|---|---|
| | $\mu E = \mu H$ | $\mu E = \mu K$ | $\mu H = \mu K$ | $\mu F = \mu I$ | $\mu F = \mu L$ | $\mu I = \mu L$ |
| Airline | 0.0000 | 0.0000 | 0.0000 | 0.0178 | 0.0000 | 0.0325 |
| USAccDeaths | 0.0000 | 0.0000 | 0.0256 | 0.0000 | 0.0000 | 0.0000 |
| WWWusage | 0.0000 | 0.0000 | 0.7147 | 0.0000 | 0.0000 | 0.0004 |
| Lynx | 0.0000 | 0.0000 | 0.0000 | 0.0000 | 0.0000 | 0.0000 |
| Nile | 0.0000 | 0.0000 | 0.0000 | 0.0000 | 0.0000 | 0.0000 |
| PetroPrice | 0.0000 | 0.0000 | 0.0000 | 0.0000 | 0.0000 | 0.0000 |
| Time series | Comparison test set | | | | | |
| | $\mu E = \mu H$ | $\mu E = \mu K$ | $\mu H = \mu K$ | $\mu F = \mu I$ | $\mu F = \mu L$ | $\mu I = \mu L$ |
| Airline | 0.0000 | 0.0000 | 0.0000 | 0.0000 | 0.0000 | 0.0000 |
| USAccDeaths | 0.0000 | 0.0000 | 0.0089 | 0.0000 | 0.0000 | 0.0001 |
| WWWusage | 0.0000 | 0.0000 | 0.0000 | 0.0070 | 0.0251 | 0.9439 |
| Lynx | 0.0000 | 0.0000 | 0.0000 | 0.0000 | 0.0000 | 0.0000 |
| Nile | 0.0000 | 0.0000 | 0.0000 | 0.0000 | 0.0000 | 0.0000 |
| PetroPrice | 0.0000 | 0.0000 | 0.0000 | 0.0000 | 0.0000 | 0.0000 |

Tabela XII

RESULTADOS DOS $p$-VALORES DOS TESTES $t$-STUDENT COM NÍVEL DE 5% DE SIGNIFICANCIA PARA A MEDIDA DE ERRO MAPE.

| Time series | Comparison training set | | | | | |
|---|---|---|---|---|---|---|
| | $\mu E = \mu H$ | $\mu E = \mu K$ | $\mu H = \mu K$ | $\mu F = \mu I$ | $\mu F = \mu L$ | $\mu I = \mu L$ |
| Airline | 0.0000 | 0.0000 | 0.0000 | 0.0171 | 0.0000 | 0.0339 |
| USAccDeaths | 0.0000 | 0.0000 | 0.0247 | 0.0000 | 0.0000 | 0.0000 |
| WWWusage | 0.0000 | 0.0000 | 0.4805 | 0.0024 | 0.0000 | 0.0007 |
| Lynx | 0.0000 | 0.0000 | 0.0000 | 0.0000 | 0.0000 | 0.0000 |
| Nile | 0.0000 | 0.0000 | 0.0000 | 0.0000 | 0.0000 | 0.0000 |
| PetroPrice | 0.0008 | 0.0000 | 0.0000 | 0.0000 | 0.0000 | 0.0000 |
| Time series | Comparison test set | | | | | |
| | $\mu E = \mu H$ | $\mu E = \mu K$ | $\mu H = \mu K$ | $\mu F = \mu I$ | $\mu F = \mu L$ | $\mu I = \mu L$ |
| Airline | 0.0000 | 0.0000 | 0.0000 | 0.0000 | 0.0000 | 0.0000 |
| USAccDeaths | 0.0000 | 0.0000 | 0.0261 | 0.0001 | 0.0000 | 0.0002 |
| WWWusage | 0.0000 | 0.0000 | 0.0000 | 0.0083 | 0.0589 | 0.7622 |
| Lynx | 0.0000 | 0.0000 | 0.0000 | 0.0000 | 0.0000 | 0.0000 |
| Nile | 0.0000 | 0.0000 | 0.0000 | 0.0000 | 0.0000 | 0.0000 |
| PetroPrice | 0.0000 | 0.0000 | 0.0000 | 0.0000 | 0.0000 | 0.0000 |